\documentclass{vgtc}

\usepackage{mathptmx}
\usepackage{graphicx}
\usepackage{times}
\usepackage{amsmath}

\title{Dynamics Based 3D Skeletal Hand Tracking}

\author{Stan Melax\thanks{e-mail:stan.k.melax@intel.com} %
\and Leonid Keselman\thanks{e-mail:leonid.m.keselman@intel.com} %
\and Sterling Orsten\thanks{e-mail:sterling.g.orsten@intel.com}}
\affiliation{\scriptsize Intel Corporation} 

\teaser{
	\includegraphics[height=1.25in]{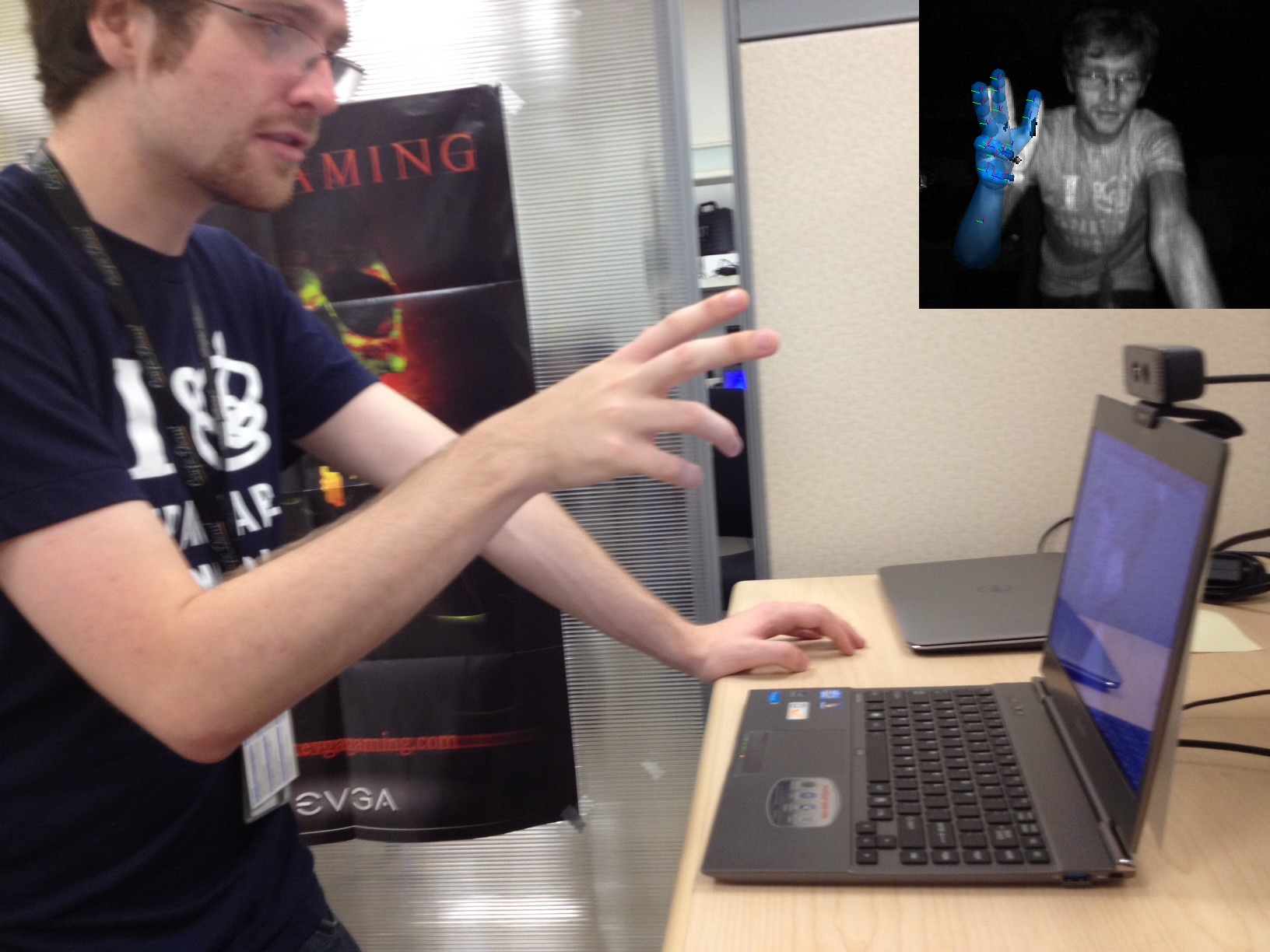}
	\includegraphics[height=1.25in]{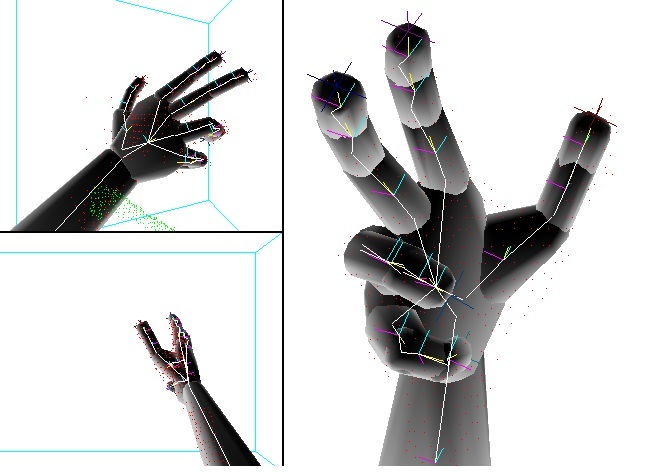}
	\includegraphics[height=1.25in]{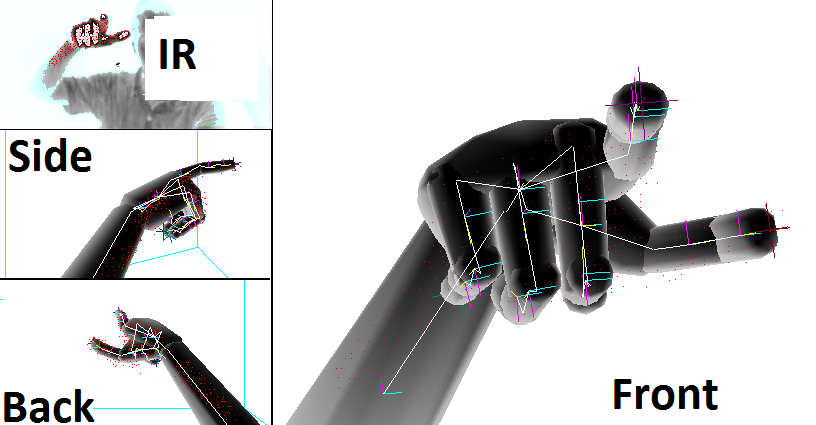}
	\caption{Hand interaction via our tracking system, along with the tracked pose of the user's hand rerendered from various viewpoints.}
}

\abstract{
Tracking the full skeletal pose of the hands and fingers is a challenging problem that has a plethora of applications for user interaction. Existing techniques either require wearable hardware, add restrictions to user pose, or require significant computation resources. This research explores a new approach to tracking hands, or any articulated model, by using an augmented rigid body simulation. This allows us to phrase 3D object tracking as a linear complementarity problem with a well-defined solution.  Based on a depth sensor's samples, the system generates constraints that limit motion orthogonal to the rigid body model's surface. These constraints, along with prior motion, collision/contact constraints, and joint mechanics, are resolved with a projected Gauss-Seidel solver. Due to camera noise properties and attachment errors, the numerous surface constraints are impulse capped to avoid overpowering mechanical constraints. To improve tracking accuracy, multiple simulations are spawned at each frame and fed a variety of heuristics, constraints and poses. A 3D error metric selects the best-fit simulation, helping the system handle challenging hand motions. Such an approach enables real-time, robust, and accurate 3D skeletal tracking of a user's hand on a variety of depth cameras, while only utilizing a single x86 CPU core for processing. 
}  

\CCScatlist{ 
  \CCScat{I.3.7}{Computer Graphics}{Scene Analysis}{Tracking};
}

\begin{document}

\firstsection{Introduction}

\maketitle

Human computer interfaces have started to take advantage of cameras and depth sensors that allow the user to naturally interact with computing devices using motions and gestures instead of a mouse, keyboard, or touchscreen.  For example, the Microsoft Kinect\texttrademark{} allows Xbox players to enjoy video games simply by moving their bodies in front of their TVs.  The input to the application is the tracked body skeleton of the torso and limbs of the user.  Extending this technology to track individual fingers in 3D is an active area of research.  Having more complete information about the user hand pose will enable richer applications including grasping, pointing, and subtle manipulation.     

Physical simulation and rigid body dynamics is a mature field of research and has become ubiquitous in professional engineering and entertainment video game applications.  Unlike narrow systems based only on kinematics (the study of motion without reference to mass or force) which are typically used to solve a specific aspect of motion,  dynamics systems incorporate momentum, forces, contacts from collision detection, joints between bodies, other external constraints on motion, and are easily extended to incorporate new data or additional requirements.  

This paper presents a computationally-efficient, camera-independent, scalable, physical-simulation-based approach for tracking 3D articulated skeletal models that is able to accurately track the human hand from a single depth sensor. Instead of using dynamics as an isolated step in the pipeline, such as the way an inverse kinematic solver would be applied only after placement of key features is somehow decided, our approach fits the hand to the depth data (or point cloud) by extending a physics system through adding additional constraints.  Consequently, fitting the sensor data, avoiding interpenetrating fingers, preserving joint ranges, and exploiting temporal coherence and momentum are all constraints computed simultaneously in a unified solver.   After a survey of similar efforts to track human motion, this paper describes our physically based tracking approach, how multiple simulations are used to increase tracking reliability, and then discuss results and applications for hand tracking.


\section{Related Work}
Hand tracking began with the VPL dataglove \cite{Zimmerman:1986:HGI:29933.275628}.  More recently, a classifier was trained to recognize the pose of a specifically colored glove as seen by a camera - a much less expensive setup \cite{Wang:2009:RTH}. Microsoft Digits avoids using special markers on the fingers themselves, instead placing a camera on the wrist \cite{KimDigits}.  To maximize consumer convenience, the eventual goal is to be able to track the hand without any apparatus \cite{visionHandPose}.

There are many works that focus on gesture recognition \cite{gesturerecognition07}.  We instead focus on the tracking, noting that gesture is then easily achieved by comparison of the relevant joint orientations from a reference pose or animation. Instead of simply providing collision geometry to a physical environment \cite{Benko:2012:MFI:2207676.2207704,Hilliges:2012:HDI:2208276.2208405}, we provide knowledge of specific hand and finger pose, with application to gesture recognition.    

There are many examples of counting extended finger by looking for extensions of the hand blob \cite{5759431}.  Typically the pixels of the hand are segmented using the depth data, and then the 2D data is used to identify features.  Such systems can be used to find fingertips.  If the 5 fingertips are properly enumerated it would be possible to use inverse kinematics \cite{ik} to provide the full pose of the human hand.  Unfortunately it is often hard to distinguish which finger is which in most cases.  Rather than separating blob detection, fingertip finding, and fitting the hand, our approach integrates all of these steps together into one system therefore exploiting all the knowledge of the mechanics of the hand, along with the depth data, to identify each finger.  

Machine learning has achieved robust full body tracking, but in some cases only up to the wrists \cite{Shotton2011}, and in constrained environments with multiple cameras\cite{Wang:2011:HMH:2047196.2047269}.  Our approach does not depend on a specific camera placement nor does it require data from multiple viewpoints.  Also, unlike many machine learning approaches, it is not trained and thus not specific to a particular camera.  It will work with different cameras (without retraining) and scale in accuracy with increased resolution, depth accuracy, and even multiple cameras. Furthermore, our approach is not specific to the human hand.  The hand is merely a 3D geometric model along with specified joint motion constraints that is provided as input to our tracking system. In many ways, our assumption of a strong geometric prior, along with a rigid body model, solved by an iterated solver to enable real-time tracking from a single range imager is most similar to recent work in full body tracking, see \cite{GooglePose}. While their approach has many technical similarities, our solution tracks a convex polyhedral model with a rigid body dynamics solver, while they use a capsule model and solve with a constrained optimization approach. Additionally, our work features a set of heuristic-driven simulations to handle the many difficult cases that arise in the case of hand tracking.

Physically motivated approaches have been applied to the reverse problem of generating realistic animations based on known user's intent \cite{Liu:2009:DMG:1576246.1531365}.  For tracking, previous work using a constrained, mechanical model of the hand showed an improved ability to track by limiting the degrees of freedom \cite{cybergloveconstraints}. Rather than limiting ourselves to a set of valid or invalid poses, we utilize range limits on joints, along with collision detection between bodies to avoid implausible hand states.  Allowing additional degrees of freedom enables more subtle and full motions of the hand \cite{accurateHandModels}.  

By leveraging depth cameras, and recursive state tracking, sequential Monte Carlo methods can search a large hand pose state-space for valid solutions \cite{particlefilterhands}. More recently, a particle swarm approach  has demonstrated the ability to fully track a temporally-coherent 3D skeletal hand model \cite{bmvc2011oikonom}.  Many sample configurations are rendered to depth buffers, which are compared to the sensor's depth data, enabling iteration toward an error minimizing pose.  In contrast to these approaches, which are very compute intensive, our approach comfortably runs over 60 Hz on a single core of an x86 processor.  At each frame we move directly toward the best local fit (pose of minimum error) using proven techniques for rigid body simulation, rather than rendering-based trial and error.  Robustness and recovering from loss-of-tracking is a significant challenge for any markerless system given the large high dimensional space of hand poses.  To increase the reliability that the system predicts the correct pose, and avoid loss of tracking, we need only spawn a few different simulations rather than testing hundreds of potential candidate poses.  Furthermore, the system is able to incorporate information, when available, such as as feature identification or known pose information acquired through other methods.

\section{Tracking Algorithm}
In general, our approach seeks to produce a stable, real-time solution to tracking a fully-defined 3D hand model. To enable this, we leverage the temporally-coherent video stream and updates a previous pose estimate with new data. Additionally, we use a strong geometric prior, specifically a convex rigid body model, to approximate the user's hand and enable efficient numerical computation. A unique technical contribution comes in our formulation of pose tracking as linear complementarity problem that handles joint constraints, surface constraints and angular constraints within a unified mathematical framework\cite{Mirtich:1995:IDS:215074.215111} that can be solved with a stable, real-time solver\cite{Catto_2005}.

\subsection{Explanation of Rigid Body Dynamics}
\begin{figure}[ht]
  \centering
  \includegraphics[width=1.0in]{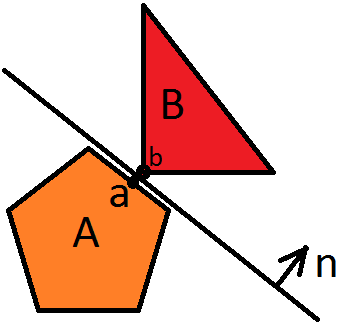}
  \includegraphics[width=1.0in]{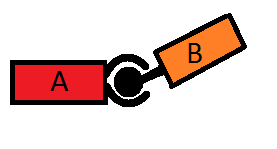}  
  \includegraphics[width=1.0in]{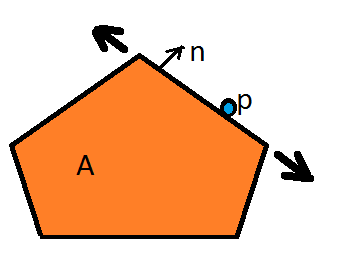}  
  \caption{Collision, Joint, and Surface constraints}
  \label{ConstraintTypes}
\end{figure}

Rigid body physics systems simulate the interaction of rigid bodies ~\cite{GDC,baraff}. Using pairwise constraints, the simulation handles both collisions between bodies and the configuration of the joints which connect the bodies. For example, interpenetration at a contact point is avoided by ensuring a positive (separating) velocity along the normal of impact of point \textbf{a} on object \textbf{A} relative to the corresponding point \textbf{b} the other object \textbf{B} (Figure \ref{ConstraintTypes}).  Similarly, joint constraints ensure zero relative velocity at the attachment point connecting the bodies.  The solver step of the physics simulation iterates a few times over all the constraints to ensure they are satisfied by applying an equal (but opposite) impulse to each body (at the point of contact and along the direction of violation) if necessary.  In the simplest of implementations, each rotational and linear degree of freedom that is limited has its own constraint associated with it.  

Relevant to our tracking system, a common feature many systems provide is the ability to limit the torque/force applied at a particular constraint.  This enables the ability to simulate objects that break under stress as well as animation blending techniques often called powered-ragdoll.  For tracking, we use constraints that exploit this cap to limit the influence when fitting to the point cloud.  We wrote our own physics simulation software, but it is very similar to the implementation of common middleware physics engines ~\cite{Bullet,PhysX}.

\subsection{Sampling from a Depth Sensor}
\begin{figure}[ht]
  \centering
  \includegraphics[width=3.0in]{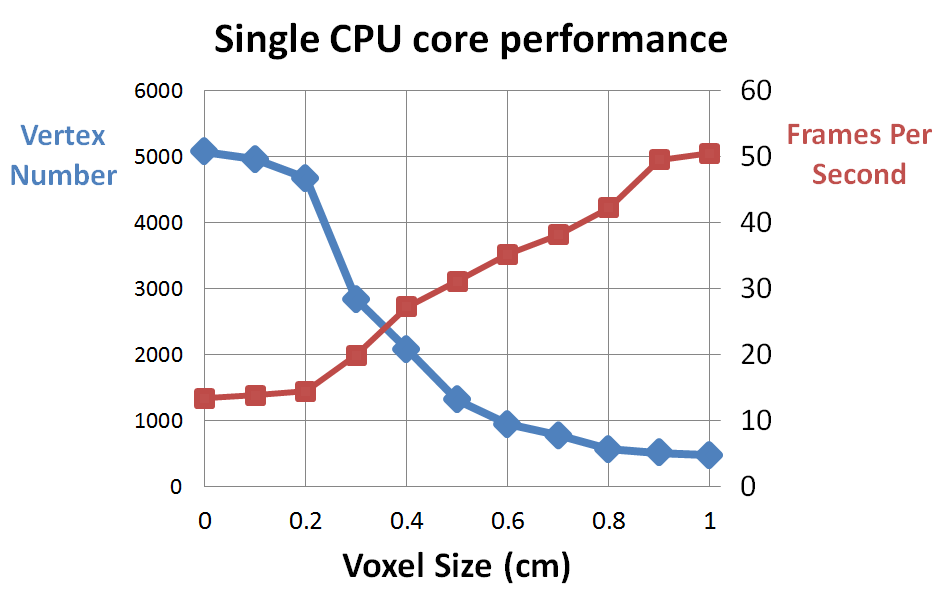}
  \caption{Subsampling Benefit. Voxel subsampling provides fine-grained control over performance by adjusting the number of constraints added to the solver. As long as the voxel size remains smaller than the size of tracked features, the algorithm is able to track correctly and the contribution of sensor noise is minimized. For adult hands, the entire voxel size range shown above produced acceptable results. Results from a 2.2GHz Intel i7 laptop. }
  \label{PerformanceGraph}  
\end{figure}

The ability to intelligently subsample the depth sensor's output has the potential to greatly improve the quality and performance of our tracking algorithm. By removing noisy camera outliers and providing a sparser, more robust data set, the tracking algorithm will converge quicker and to a more accurate result. For this work, we implement a voxel grid subsample, which has shown success in the space of real-time reconstruction \cite{modelreconstrunction}. For each occupied voxel, we generate a single point, positioned at the center-of-mass of the points in the voxel. This technique is fast, removes camera noise, guarantees uniform sampling across the volume, and can be used for fine-grained control of performance, as seen in Figure \ref{PerformanceGraph}. However, selecting a candidate point cloud volume from the range-imager is outside the scope of this work; in practice, we track the closest, sufficiently large volume in the scene. 

\subsection{Point Cloud Samples as Constraints}
\begin{figure}[ht]
  \centering
  \includegraphics[height=1.25in]{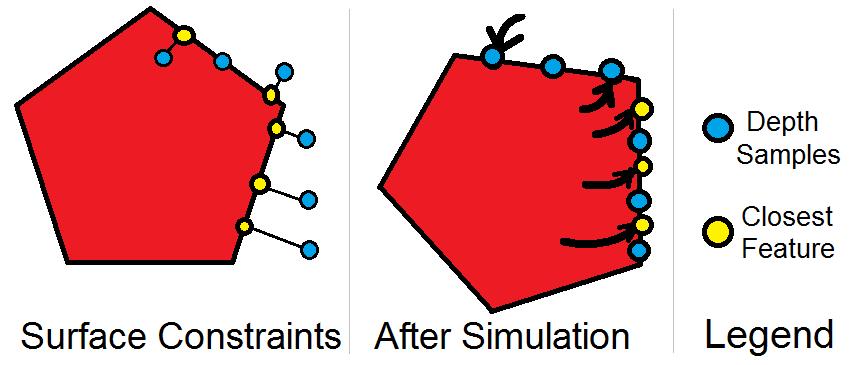}
  \caption{Creating Surface constraints by Mapping 3D Points To Facets on a Body}
  \label{fig:magnets}
\end{figure}

Using physics simulation for tracking begins with a 3D model of the real-world object being tracked and then using the depth sensor data to move the model.  The data from a depth sensor is a rectangular grid of pixels indicating distance to closest object from the device along each ray.  This provides a collection of points in 3D indicating geometry that is viewable from the camera.  Like collisions and joints, we use each of these points as a magnet-like constraint.  In particular, we constrain the camera facing surface of our 3D tracking model to these points.  The position in 3D space is known, but the position on the model (or even which rigid body) is unknown.   As a heuristic we use the closest feature (face) for attachment.  These attachment constraints are one dimensional, similar to the approach used to resolve object interpenetration in modern physics engines. In these systems, objects which have come to overlap are constrained to move away from one another along their surface normals, without restricting parallel motion.  In our approach, rigid bodies which are physically separated from the point cloud are constrained to bring them \emph{into} contact, while retaining their ability to move laterally to seek a better fit, as shown in figure~\ref{fig:magnets}.

For tracking a model with multiple rigid bodies, we first find the rigid body closest to point cloud samples and then create a surface constraint on the nearest feature.  
Clearly, surface constraints provide the same functionality as Iterative Closest Point \cite{ZHANG92iterativepoint}, but now formulated within a rigid body system which allows for articulated models (multiple bodies) and other constraints (joints and contacts).

\subsection{Impulse Cap}
Clearly this approach assumes temporal coherence in that the system begins with a correct initial fit and that it is able to continuously update each frame.  Even under continuous motion, it is not possible to ensure that all the depth points will be mapped to the correct spot on the model.  There are hundreds of points in the point cloud and usually most of them are placed correctly.  Therefore we limit the amount of impulse (force/torque) that each surface constraint can apply at the solver step.  Specifically, we choose a maximum that is just strong enough to pull the hand when the user is engaged in very fast motion.  The constraints used for attaching the bones and limiting finger motion have no impulse cap.  We assume that under normal interactive usage, the user's hand will remain intact.  
In contrast to spring based penalty methods \cite{Moore:1988:CDR:54852.378528} or other methods that would minimize root mean square error, 
constraint-based iterative impulse techniques \cite{Erleben:2005:PA:1051390} solve overconstrained systems by overriding 
the influence of outliers and fitting to the majority.  By limiting the constraints generated by depth values, erroneously 
created surface constraints are quickly overpowered by all the other constraints in the system.
As a result, the bones click into place like puzzle pieces or there is a clear indication that the system has failed to find a fit.

\subsection{Boundary Planes}
In addition to telling us where things are, the sensor depth data also indicates free space, places where something cannot be.  The region between the device and each depth point is known to be free of geometry.  We take advantage of this by adding boundary collision planes, taken from the hull of the point cloud, to contain the 3D model.  Because small objects, such as fingers, can move very fast relative to their size, we do not add blocking collision planes within this convex envelope since we want bodies to be able to freely move within this space toward the best fit.  

\subsection{The Hand Model}
For short range user tracking, we use a 3D model of the user's hand.  Although the human hand has soft deformable regions, it is possible to model and rig a rigid-body representation that adequately approximates a user's hand geometry and motion that has less error than what is coming from the depth sensors.  Each rigid section of the wrist and hand is approximated by a convex rigid body.  Although the palm consists of a number of anatomical bones, it is only 1 rigid body in our system.  Fingers and thumb each consist of 3 rigid bodies each.  Just as human bones are connected and motion limited with ligaments, the virtual model is specified with linear attachments and angular limit constraints to prevent physically impossible configurations but still allow for abduction and adduction.  Our reference hand model is 20 cm long, but is easily scalable at runtime to fit the user's hand length and width to help tracking accuracy.  Furthermore, the model is just a data file, created in and exported from 3DS Max, that is easily modified to allow our system to work with an individual with a unique hand such as double jointed mobility or missing digits. Our standard hand size works well for a variety of users, and morphological surveys of the human hand lend credence to the fact adult hands are often of similar size and proportion \cite{handSizeMeasurements}. However, while we do not focus on this work, there do exist techniques for automatic generation of accurate 3D hand models based on reference images \cite{Rhee:2006:HHM:1111411.1111417}.

\begin{figure}[ht]
  \centering
  \includegraphics[height=0.85in]{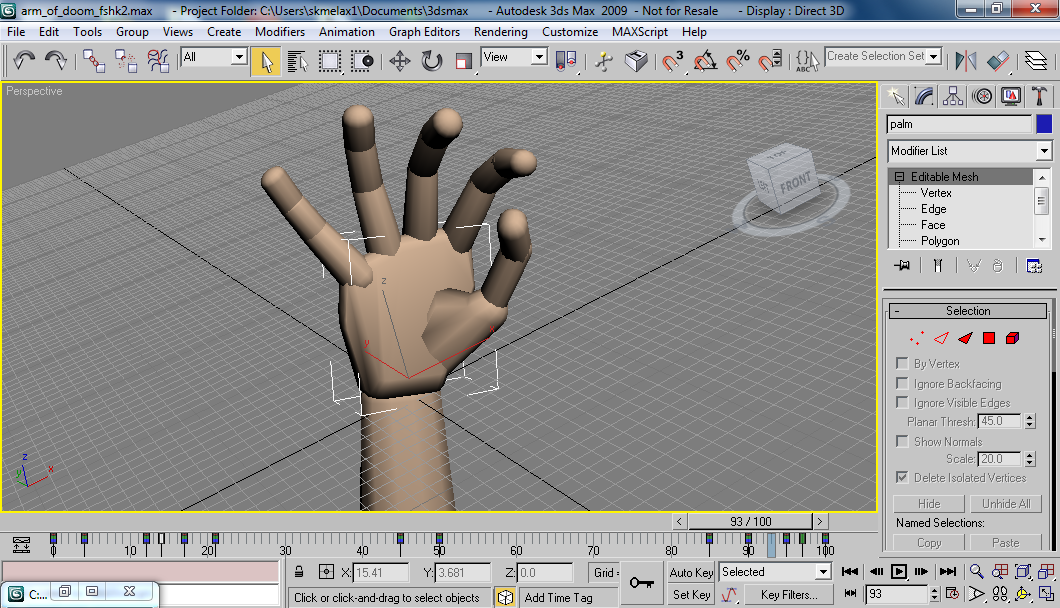}
  \includegraphics[height=0.85in]{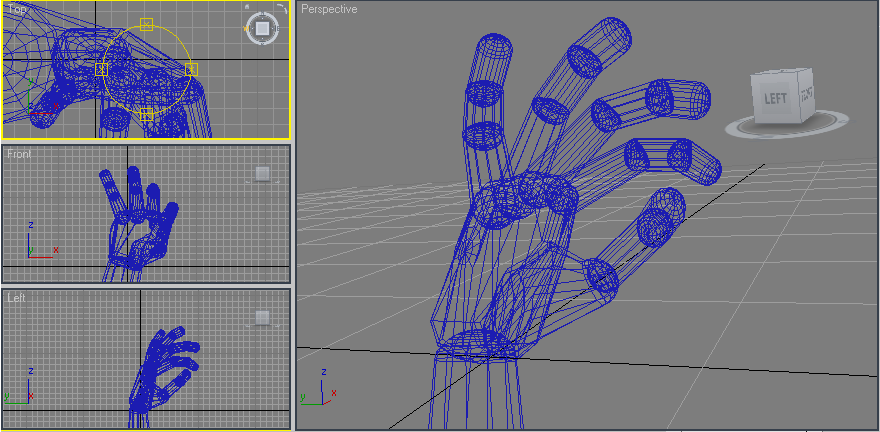}  
  \caption{Hand Modeled in 3DS Max. Wireframe mode shows Interpenetrating bones.}
\end{figure}

Our model is constructed to help minimize errors in the system. The geometry of the rigid bodies interpenetrates the neighboring bones to avoid gaps. For example, the finger bones are roughly capsule shaped with the rounded caps occupying the same space as its neighbor.  Collisions between neighboring shapes is disabled.  We want the tracking surface constraints, described below, to only limit motion parallel to the outer surface and to be able to slip from one rigid body to the next.  Rather than true capsules, we used convex polyhedra to better fit the geometry of the actual hand.

\begin{figure}[ht]
  \centering
  \includegraphics[width=3in]{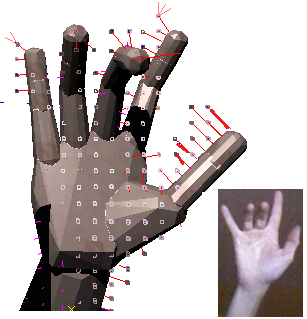}
  \caption{Surface constraints from Point Cloud (Depth Values) Affecting the Bones of the Hand}
\end{figure}

\subsection{Initializing and Exploiting Higher Level Information}
Normally we initially know little about what object is at a particular depth value.  Occasionally, the user will place his hand in an obvious pose such that a simple computer vision technique can recognize the state with a high degree of certainty.  Our physical simulation system is easily able to exploit this higher level information, when available, by constraining these points to their known areas on the tracking model (rather than just the closest).  Furthermore, even partial information is still useful; for example, if we know a point is on a fingertip, but don't know which one, then we limit that point's matching to just the fingertips.  

\section{Multiple Simulations and Error}
Occlusions and camera limitations make it necessary to exploit temporal coherence, yet also means that we cannot guarantee correctness either.  Following a point cloud works best when starting from the correct state and the changes are small.  Sometimes the system gets trapped in a local minimum where the state is incorrect but moving to the correct solution is not just a matter of updating rigid bodies closer to the nearby point cloud. Another issue is that not all the higher-level information fed into the system has 100\% confidence.  For example, a computer vision pre-pass on a particular frame might mistakenly classify that the thumb is raised when in fact the index finger is extended.  We need to be able to deal with all these possibilities.  Fortunately, even with all our added constraints, physical simulation of a 17 bone hand is a light computational load.  We can perform many of these per 17 millisecond frame.  Therefore, our system spawns a handful of simulations - each incorporating a different subset of possible assumptions.  The tracking system evaluates the resulting pose from each simulation and returns the one with the least error or best fit with the depth data.  The different strategies used by the various simulations are motivated by our observations of common user motions, especially those that lead to difficult or ambiguous depth samples. With multiple heuristic-driven models searching the state-space, each one provides a lower-bound on a certain behavior set, and together they are able to limit degenerate behavior of the tracking system. 
\begin{figure}[ht]
  \centering
  \includegraphics[width=3.0in]{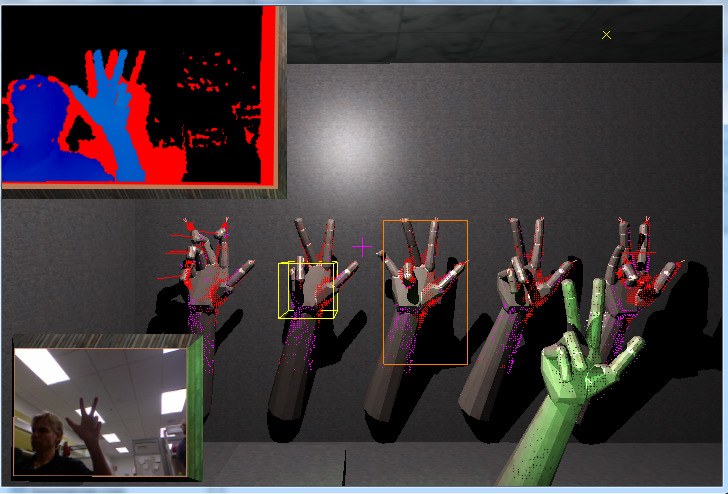}
  \caption{Multiple simulations exploring more possibilities.  Best fit shown in front.}
\end{figure}

\subsection{Gross Motion Bias}
Hand motions can be very fast relative to its scale.  In a single frame, a user's fingers can easily move a distance greater than their width. It is easy to imagine a situation where all the fingers are shifted one digit over when comparing the next frame's point cloud to where the model finished the previous frame.  Therefore, we have a dedicated simulation where the entire hand is first frozen into a single rigid body - disallowing relative motion between bones.   The constraints aligning the palm and the wrist, along with the envelope of collision planes on the boundary, overcome point cloud samples mapped to the wrong finger and moves the model closer to where it should be.   Then the hand model is unfrozen, the point cloud is re-mapped, and a regular simulation takes place to update any local movements.  An additional benefit of first moving the full model in unison is that this provides a logical positional update for any occluded parts.
\begin{figure}[ht]
  \centering
  \includegraphics[width=3.0in]{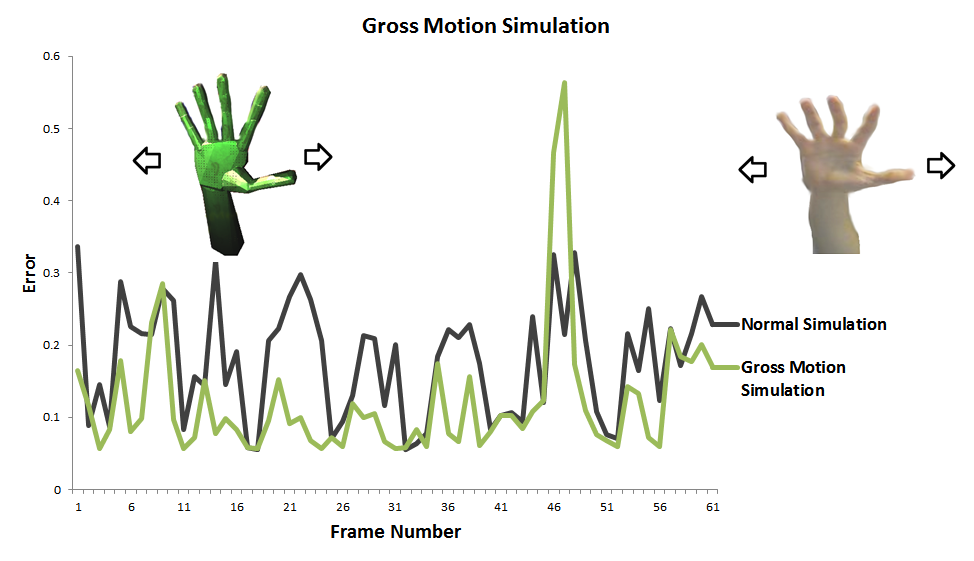}
  \caption{Gross motion biased simulation showing its better fit over a 2 second long waving motion. While the normal simulation exhibits fitting error, the frozen hand simulation correctly fits the large-scale motions of the hand.}
\end{figure}

\begin{figure}[ht]
  \centering
  \includegraphics[width=3.0in]{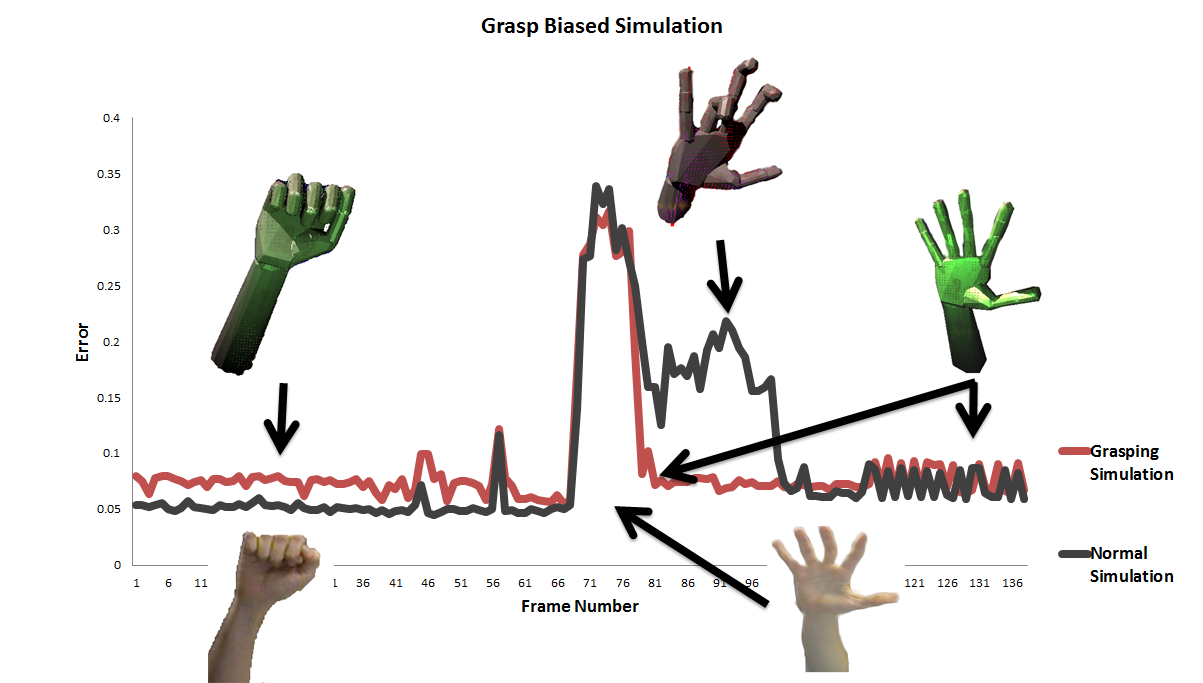}
  \caption{Grasping biased simulation showing a faster response to the user opening their hand. While running simulations independently (not sharing pose information), a user rapidly changes from a closed fist to an open hand.  The grasping simulation quickly moves all finger together into the correct position, while the normal simulation takes more frames to converge to the correct fit.}
\end{figure}

\subsection{Grasping Bias}
Often, users will move their fingers in unison, whether they are all curled in a fist, laid out flat or somewhere in between.  Since it is so common, we have a dedicated simulation where the hand model has additional orientation constraints across the knuckles to ensure the fingertip bones are within 10 degrees of being parallel to their neighbors and likewise for the finger mid bones.  The result is a simulation that is very good at tracking grasping motions reliably and helps to properly place occluded digits.  When the user is pointing, the simulation output is typically rejected due to measured error.   

\subsection{Searching Nearby States}
Sometimes the system gets trapped in a local minimum.  For example, a user may be pointing with his index finger, but the system is fitting the thumb or middle finger to the extended point cloud data instead.  In other cases, it is possible that the wrong numbers of fingers are raised and local attraction is not going to raise or lower the extra finger. Perturbation of individual degrees of freedom has been used to explore nearby hand model states \cite{accurateModelTracking}. In our system, another simulation begins with the best fit pose from the previous frame and adds constraints to raise or lower a finger into the opposite pose of where the system currently thinks it is.  Sometimes it also grabs the neighboring finger and flips that one as well.  Because it may take a few frames to settle the entire hand to best fit the depth data, the choice of which fingers (and how many) to flip cycles every half second.  Given good depth accuracy and model matching the user's hand, the system is able to on-the-fly correct these sorts of mistakes in the tracking.  
\begin{figure}[ht]
  \centering
  \includegraphics[width=3.0in]{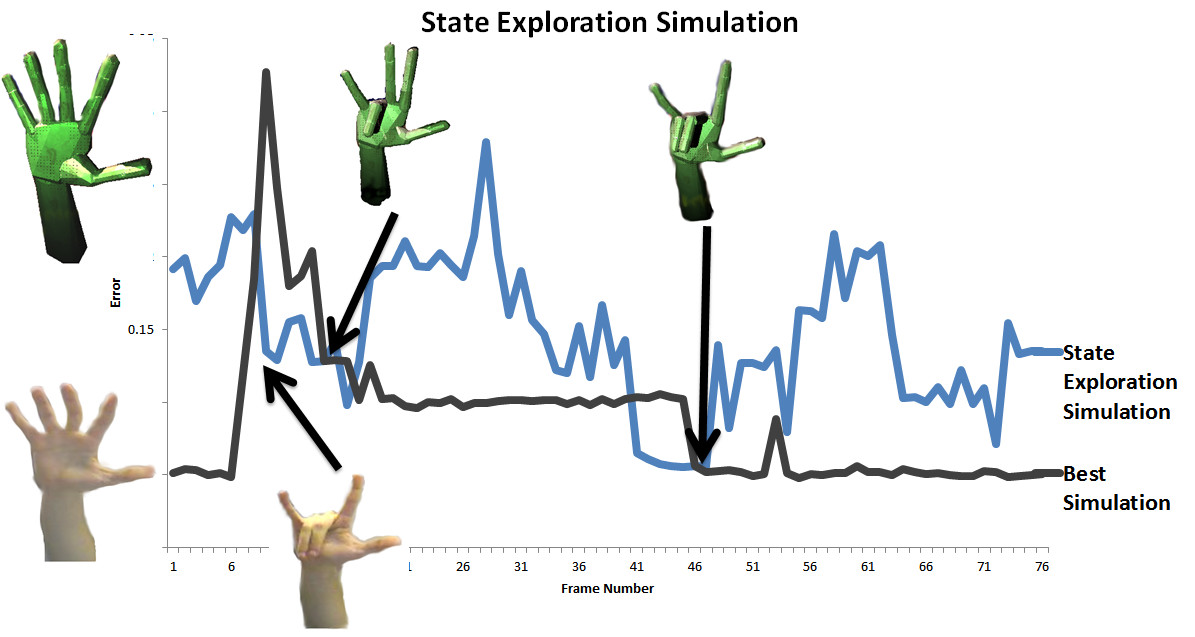}
  \caption{Finger flipping simulation correcting the model's fit after rapid motion. As the user quickly retracts both their middle and ring finger, the finger flipper attempts a prospective candidate pose with the ring finger retracted, which exhibits better fit and gets adopted. Soon after, the finger flipper simulates a candidate with the middle finger bent over, which is also adopted as a better fit.}
\end{figure}

\subsection{Feature Detection}
Image based pose classifiers and feature detection systems can generate false positives.  When applied to the physics based tracking, we found that these systems can throw off the tracking as often as they initiate or regain lost tracking.  With multiple simulation candidates we were able to dedicate a specific simulation that incorporates any labeled appendages so that we only incorporate this high-level data into the final output when it does, in fact, lead to a better result.  

In particular, one of the simulations in our tracking monitors a naive five finger detector. The implementation checks scanlines for contiguous segments and triggers when five segments come together; the top of each segment is then attached to a fingertip and a physics simulation is run. Obviously such a detector is not very sophisticated, but it is sufficient to trigger pose resets when the open hand is visible and vertical. As mentioned, this prospective simulation is typically discarded by our 3D error metric in cases when it is incorrect. This approach for selective binding can be supplemented with far richer finger detectors \cite{Malik03real-timehand}, but they are not yet explored in this work. 

\subsection{Error Metric}
Rather than using depth buffer comparisons, we try to measure the error in 3D. The most logical error metric would measure the distance between a set of points on the virtual hand model, and corresponding points on the user's hand. Unfortunately, the point cloud data is markerless, so we do not know what the corresponding points are. 

We do know that if there is a point in the point cloud $P$ a given distance away from the surface of the tracking model $B$, then the model must be moved at least that distance to match the actual state of the user's hand. In practice, we find it useful to define an error $errFit_i$ for each rigid body $B_i$ in the model $B$, based on the minimum distance that body would have to be displaced to fit the point cloud data. We define this as the distance of the furthest point $p$ in the set of point cloud points $P_i$ which are as closest to the surface of $B_i$ (the current rigid body) as they are to any point on the surface of the hand, as shown in equation ~\ref{eq:fit}. 
\providecommand{\abs}[1]{\lvert#1\rvert} \providecommand{\norm}[1]{\lVert#1\rVert}
\begin{equation}
P_i = \{p  :  p \in P \text{, } \min\{\abs{p-b_i} :  b_i \in B_i\} = \min\{\abs{p-b} : b \in B\} \} \label {eq:p_i} \\
\end{equation}
\begin{equation}
errFit_i = \max \{ \min\{ \lvert p-b_i \rvert : b_i \in B_i\} : p \in P_i\} \label {eq:fit}
\end{equation}
\begin{equation}
 errOcc_i =
  \begin{cases}
   \text{r} & \text{if } centroid(B_i) \text{ occludes background}\\
   0       & \text{otherwise}
  \end{cases} \label {eq:occ}
\end{equation}
\begin{equation}
errTotal = \sum{errFit_i} + \sum{errOcc_i} \label {eq:total} 
\end{equation}

In addition to generating error from points far from the tracking model, we also penalize rigid bodies which occupy space where the depth camera indicates nothing is present. Ideally we would want to take the distance the object would need to travel (either sideways or toward the back) in order to not be in front of the point cloud.  Since this would require costly searching for the optimal direction of displacement we instead apply a constant penalty.  In particular we take the centroid of each bone, which, if not occluded by pixels of the depth image has an error at least as large as the radius $r$ of the largest inscribed sphere within the bone. This is seen in equation \ref{eq:occ}.

The overall error metric for a hypothetical pose is therefore computed by summing up the per body errors and the penalty terms (eq. \ref{eq:total}). Unlike depth buffer comparisons, the metric is not symmetrical, however it is less sensitive to the numbers of pixels and it does indicate when there is a fit. This combination of fitting (eq. \ref{eq:fit}) and occlusion (eq. \ref{eq:occ}) error penalizes both local misalignment and obstruction of background in a similar fashion as the error metric proposed in \cite{GooglePose}.  This error is not used to adjust the model's pose -- only to measure the outcome and pick the best fit. 

\section{Results and Discussion}

There is currently no standard evaluation of the robustness of tracking solutions.  The high fidelity of our hand tracking solution is demonstrated in the video accompanying this paper.  We describe the successes and challenges of our system on a variety of motions on different camera setups.   

\subsection{Stereoscopic Camera} 
We have tested our algorithm on a Point Grey Research Bumblebee XB3, along with OpenCV's block-matching implementation, to generate real-time depth. Since our algorithm does not require contiguous blobs or points, we were able to achieve reasonable tracking despite many missing pixels and incorrect, noisy samples. All of these factors affected the quality of our tracking and in the best of circumstances it was only able to track slow, careful motions. Although this implementation was academically interesting, it was certainly not sufficient for usage in an application.

\subsection{Structured Light}
The ASUS\textregistered{} Xtion PRO camera provides depth data using structured light.  The data is fairly accurate, but sometimes there is missing data.  Small features, such as fingertips facing the camera or child sized fingers, often do not even show up.  Unfortunately, this can not be solved by moving closer to the camera since the camera was designed for full-body interaction and has a minimum distance of around 0.6m.  The tracking is surprisingly tolerant of missing data, and often retains the correct pose if only a few frames of data are lost.  As expected, the tracked hand is not as smooth as the user's motions.  Tracking is sometimes lost, requiring the user to put his hand in an easily identifiable pose or otherwise move until tracking is regained.  Our hand tracking, for adult hands, does work with this device up to its limits - not fully robust, but on par or better than other markerless solutions.  

\begin{figure}[ht]
  \centering
  \includegraphics[height=1.75in]{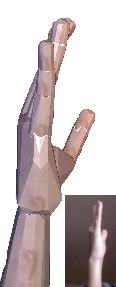}
  \includegraphics[height=1.75in]{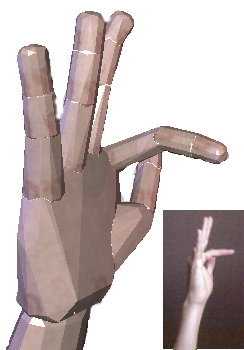}
  \includegraphics[height=1.75in]{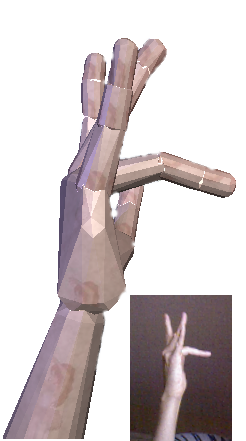}
  \caption{Starting from a knife hand gesture with heel of the palm toward camera and then bending fingers initially occluded, the surface constraints pull down the correct digit.}
\end{figure}

\subsection{Time-of-flight Camera}
\begin{figure}[ht]
  \centering
  \includegraphics[width=3.0in]{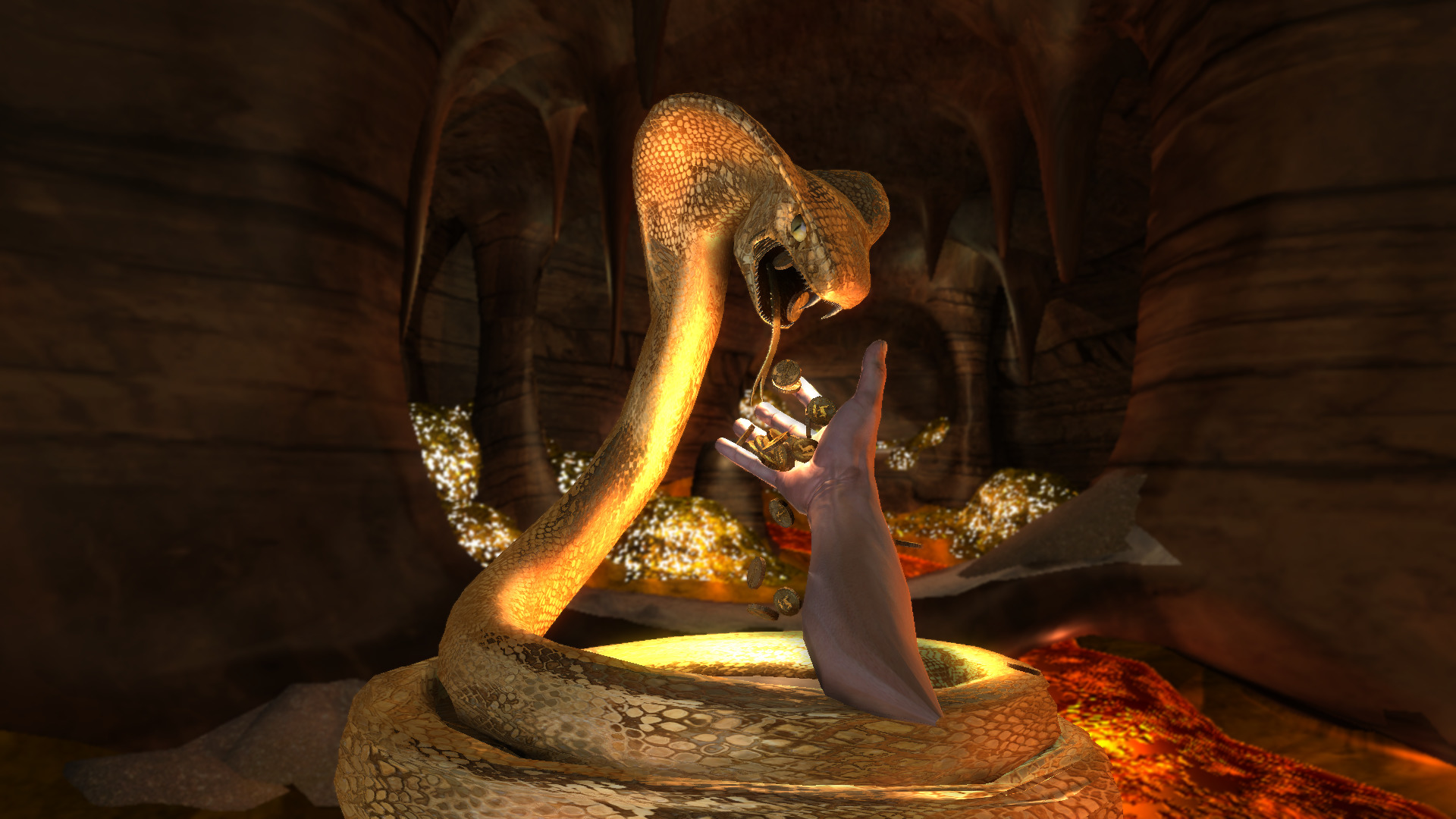}
  \caption{ Demo at the Consumer Electronics Show 2013 }
  \label{vomit}
\end{figure}

Creative's Interactive Gesture Camera\texttrademark{} is a small portable device designed to work with short-range usages and has an effective range starting at 20cm.  Accessible through the Intel\textregistered{}  Perceptual Computing SDK \cite{percSDK}, it provides direct access to the depth data, which is generated using time-of-flight technology. Since the hand can be placed close to the camera, it is able to generate high-fidelity data. Fingers pointed at the device are not lost and this enables continuous updates, with fewer losses in tracking as the hand motion passes through these circumstances. Although the data is sometimes noisy, the SDK provides a fast smoothing algorithm that produces reliable data. Additionally, we've obtained good results by simply using a median or bilateral filter \cite{fastmedianfiltering}. There are still situations, most notably very fast motion, that can still throw off the tracking. Additionally, rolling a clenched fist often fails, as the depth samples are too similar to induce the required rotation.  The system was demonstrated at Intel's 2013 CES booth with unmanaged public usage and indicated a usable quality of tracking (Figure \ref{vomit}). 


\subsection{Projected Texture Stereoscopic Camera}
The best results are seen from preliminary testing on the TYZX DeepSea G3 Vision System, a precursor to the Intel RealSense R200 \cite{2017arXiv170505548K}, which emits a light pattern and uses two IR sensors to construct a depth image. Running at 60fps, and having a minimum distance of 20cm, it has high resolution and accuracy, with low camera noise and few missing pixels.  Not surprisingly, this results in more precise tracking, which should enable faster hand motion and more fine grain manipulation within applications.  Even with this setup, the tracking can get lost, requiring the user to put his hand in an easily identifiable pose to regain the tracking.  Fortunately, with the improved precision, measuring error and identifying the correct pose amongst multiple candidates improves greatly.  

\subsection{Tracking of Multiple Hands}
Our tracking system can be easily extended to track both of the user's hands. There are many possible approaches to segmenting the hands \cite{Moeslund:2001:SCV:376890.376905}, and we opted for a computationally efficient solution. Given that the user's hands are the foremost objects in the scene, we use a k-means clustering algorithm \cite{Kanungo:2002:EKC:628329.628801}, to cluster the point cloud into two hand clusters; we assume the left cluster to be the left hand, and the right cluster to be the right hand. If the clusters are too close, we merge them into a single cluster and use it to track the right hand. Additionally, this approach preserves the 3D sample-based nature of our tracking algorithm.

\section{Application}

Today, there already exists a wealth of interaction usages with 3D hand position data \cite{LaViola:2011:SIA:2037636.2037637}.  Rich interactive applications can be built merely by assembling physically enabled content \cite{rocket}.  Such physics scenes are ideal for adding hand interaction.  With the skeletal pose information from a tracked hand, it is straightforward to use these to animate the bones of a 3D model such as the rig of a skinned hand model.  Such a hand model can then use collision detection to interact with virtual objects \cite{dataglovegrasping,Lam:2004:MED:1067343.1067393}.  Without real-world or haptic feedback, when the bones are controlled directly in this manner it can result in a hand that is too strong.  There is nothing preventing it push a heavy box at high speed or penetrating right through a wall.  Therefore we use the relative bone orientations from connected bones in the tracking data to drive the ``muscles'' of a virtual hand in the application - often described as the ``powered ragdoll'' approach \cite{GDC}.  The interactive simulation system in the application applies forces and torques (up to realistic limits) in an attempt to have the virtual hand match the hand pose provided by the tracking.  Consequently when the user's real hand makes a fist while there is an object in front of the palm of the 3D hand model, the virtual fingers will wrap tightly around the object and apply forces to maintain the grasp.  

\begin{figure}[ht]
  \centering
  \includegraphics[width=3.0in]{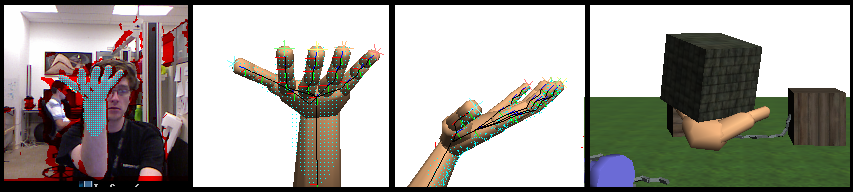}
  \caption{Picking up a block in a physically enabled virtual environment.}
\end{figure}

\begin{figure}[ht]
  \centering
  \includegraphics[width=3.0in]{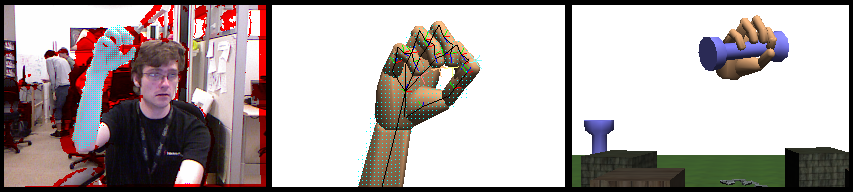}
  \caption{Interacting with a barbell where geometry and collision (instead of gestures and custom programming) enables the grasping, holding, and manipulating of the object.}
  \label{barbell}
\end{figure}

\begin{figure}[ht]
  \centering
  \includegraphics[width=2.0in]{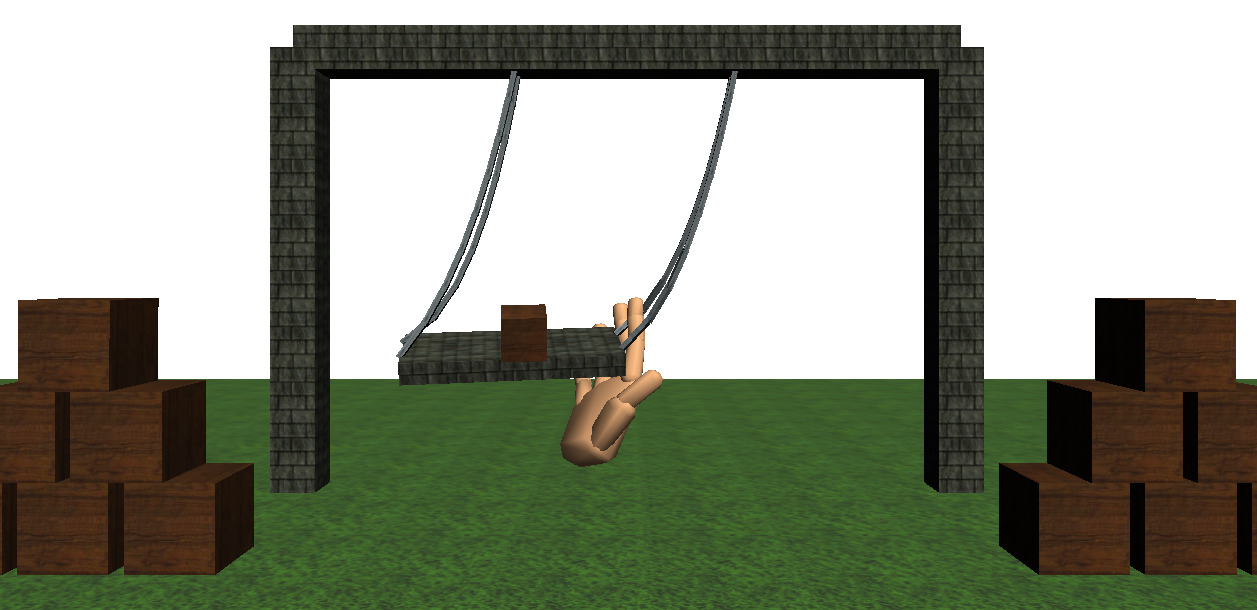}
  \includegraphics[width=1.0in]{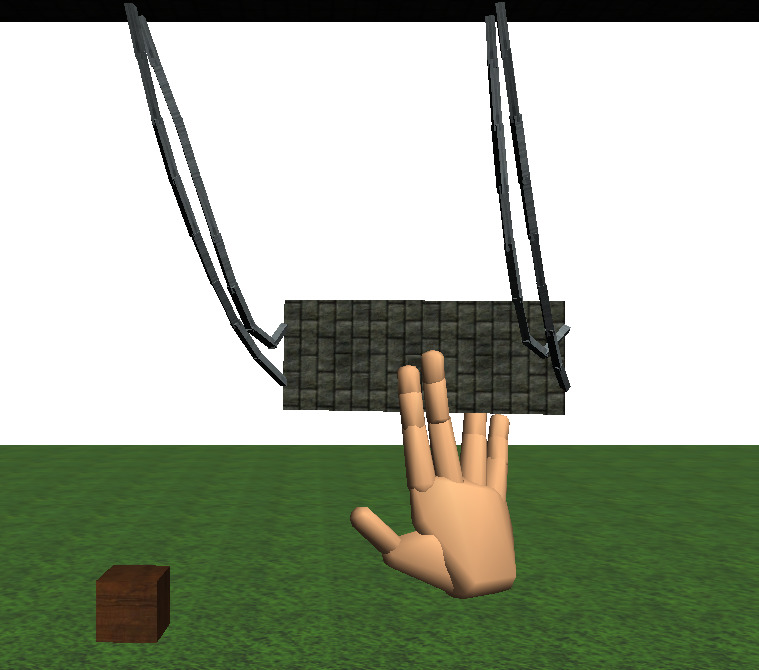}
  \caption{Interacting with highly jointed objects.}
  \label{swingset}
\end{figure}

\begin{figure}[ht]
  \centering
  \includegraphics[width=3.0in]{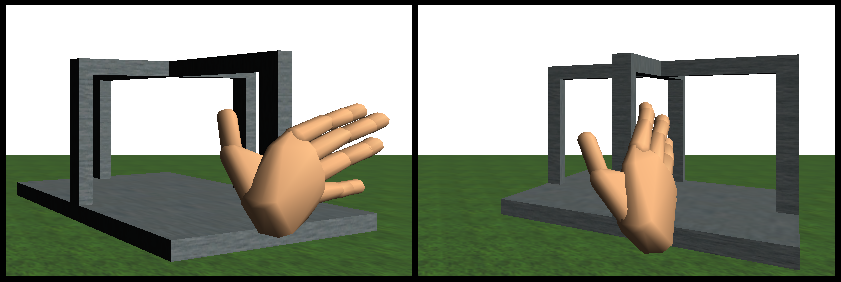}
  \caption{Interacting with object with 1 rotational degree of freedom.}
  \label{merrygosquare}
\end{figure}

Our preliminary results testing such systems found that it can sometimes be challenging to manipulating rigid-body objects.   We found a number of things that improved the user experience.
\begin{itemize}
\item Instead of using the model from the tracking system, its better to use a virtual hand shaped better for the tasks in the application.  Its easier to grasp with a deeper or concave palm rather than the enlarged convex palm of the tracking model.  
\item Similarly, objects become easier to manipulate when they are constructed in a way that interact well with the virtual hand model.  Unlike a convex polyhedron, a barbell that is capped at either end with larger geometry will prevent it from slipping out of the virtual hand (Figure \ref{barbell}).
\item Compared to convex bodies with 6 degrees of freedom, users have greater success with partially constrained objects.  For example, a throw switch/lever or a dial are examples of 3D objects that interact reliably (Figure \ref{merrygosquare}).  
\item When the user moves his hand outside the field of view of the depth sensor tracking is lost.  This can be overcome by translating and/or rotating the user view in the 3D application as the hand nears the sensor's limits.  This not only encourages the user to keep his/her hand in the proper space, it lets him/her navigate about the scene.
\end{itemize}
\newpage
\section{Conclusion}

We have presented a system for fully articulated 3D tracking the human hand from markerless depth sensor data. The system is able to integrate information about the object being tracked and fully utilize the 3D depth data collected at runtime.  The tracking fidelity improves with increasing resolution, better model accuracy, and faster camera frame rate.  The system runs comfortably on a single CPU core, but, using additional computing resources, could be extended to search and simulate more possible poses in order to improve robustness further. 

\section{Future Work}

Many application concepts, such as accurately playing an invisible instrument or American sign language, will require much more fidelity and robustness than what is possible today.  Fortunately, our tracking system has potential for improvement.  Currently our 3D hand model used for tracking consists only of convex collision bones used to fit to the depth data or point cloud.  We would like to to add softbody support \cite{softbody} to enable a more geometrically accurate representation of the human hand.  We anticipate this will add more accuracy especially as the resolution and accuracy of depth sensors improves in the future.   

In early experiments, using primary colors on the fingertips resulted in flawless tracking.  Other than this test, we have not explored the potential of using the RGB data to assist the tracking.   A system that would look for and track natural markers or color features could add value to the tracking.  If IR interference is not a problem, multiple sensors could also provide better coverage.

So far the focus has mostly been on tracking a single hand.  The current system can potentially be used to track any model such as a full body skeleton.  We anticipate additional work would be required to meet performance requirements and make this adequately robust.  For example, a multiresolution approach, similar to what we currently do for fast hand motions, might be most effective when mixing small and large bones (torso sections and finger bones) into a single model.  

As with many emerging technologies, a wide open research problem is finding applications and usages that take advantage of the new capabilities that are offered. Although we haven't explored such examples in this work, we note that a real-time 3D hand tracking system can theoretically be used in place of any existing 2D multitouch interface, with the exclusion of haptic feedback. This would allow of application of this work from world-in-miniature interactions \cite{Coffey:2011:SWM:1944745.1944777}, through playing virtual instruments \cite{Ren:2012:TET:2159616.2159618}.

\acknowledgements{
The authors wish to thank Achin Bhowmik and Mooly Eden for starting Intel Corporation's Perceptual Computing project and supporting this research. }
\newpage

\bibliographystyle{abbrv}
\bibliography{template}
\end{document}